%% file: main.tex
\documentclass[10pt,twocolumn,letterpaper]{article}

\usepackage[pagenumbers]{iccv}             % To produce the CAMERA-READY version
\usepackage{mdframed}
\definecolor{iccvblue}{rgb}{0.21,0.49,0.74}
\usepackage[pagebackref,breaklinks,colorlinks,allcolors=iccvblue]{hyperref}
\usepackage{multirow} 
\def\paperID{5390} % *** Enter the Paper ID here
\def\confName{ICCV}
\def\confYear{2025}

\title{BiPVL-Seg: Bidirectional Progressive Vision-Language Fusion with Global-Local Alignment for Medical Image Segmentation}  

\author{
Rafi Ibn Sultan$^{1}$, \quad Hui Zhu$^{1}$, \quad Chengyin Li$^{1}$,  \quad Dongxiao Zhu$^{1}$\thanks{Corresponding author.}\\ 
 $^{1}$Wayne State University\\
{\tt \small \{rafisultan, hui, cyli, dzhu\}@wayne.edu} \\
}

\begin{document}
\maketitle
\input{sec/0_abstract}

\input{sec/1_intro}

\input{sec/2_related_work}

\input{sec/3_method}

\input{sec/4_experiments}

\input{sec/5_conclusion}
{
    \small
    \bibliographystyle{ieeenat_fullname}
    \bibliography{main}
}

\input{sec/X_suppl}

\end{document}

%% file: sec/0_abstract.tex
\begin{abstract}
Medical image segmentation typically relies solely on visual data, overlooking the rich textual information clinicians use for diagnosis. Vision-language models attempt to bridge this gap, but existing approaches often process visual and textual features independently, resulting in weak cross-modal alignment. Simple fusion techniques fail due to the inherent differences between spatial visual features and sequential text embeddings. Additionally, medical terminology deviates from general language, limiting the effectiveness of off-the-shelf text encoders and further hindering vision-language alignment. We propose \textbf{BiPVL-Seg}, an end-to-end framework that integrates vision-language fusion and embedding alignment through architectural and training innovations, where both components reinforce each other to enhance medical image segmentation. BiPVL-Seg introduces bidirectional progressive fusion in the architecture, which facilitates stage-wise information exchange between vision and text encoders. Additionally, it incorporates global-local contrastive alignment, a training objective that enhances the text encoder’s comprehension by aligning text and vision embeddings at both class and concept levels. Extensive experiments on diverse medical imaging benchmarks across CT and MR modalities demonstrate BiPVL-Seg’s superior performance when compared with state-of-the-art methods in complex multi-class segmentation. Source code is available in this \href{https://github.com/rafiibnsultan/BiPVL-Seg}{GitHub Repository}.
\end{abstract}

%% file: sec/1_intro.tex
\section{Introduction}
\label{sec:intro}

Medical image segmentation traditionally relies on training models using images paired with corresponding ground truth masks~\cite{ronneberger2015unet,hatamizadeh2021swin,li2023focalunetr,li2023autoprosam}. Despite increasing interest in multimodal methods, single-modality learning remains the dominant approach. However, this overlooks how clinicians operate in real-world practice, where both visual data and textual reports are jointly considered for diagnosis, and this combined information directly influences segmentation decisions~\cite{qin2022medical, oh2023llm, zhao2023one}. To better reflect clinical workflows, there is growing interest in Vision-Language Models (VLMs)~\cite{radford2021learning,luddecke2022image} for medical image segmentation. By integrating complementary visual and textual information, these models aim to improve segmentation performance~\cite{oh2023llm,huang2024cat,liu2023clip,li2024novel}.

\begin{figure}[t]
\centering
\includegraphics[width=0.9\columnwidth]{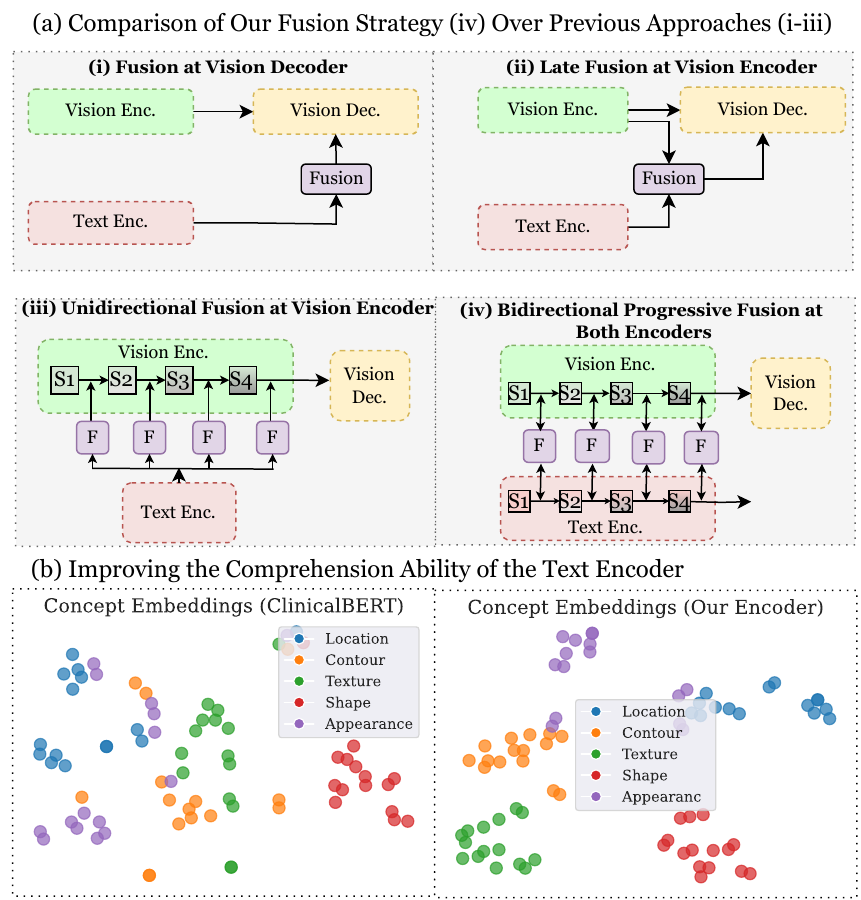}
\vspace{-10pt}
\caption{(a) Comparison of our bidirectional progressive fusion (iv) with prior VLM architectures (i–iii), showing improved cross-modal interaction. (b) t-SNE visualization shows an improved concept separation in BiPVL-Seg’s text encoder (right) compared to ClinicalBERT (left), demonstrating the benefit of global-local alignment.}
\vspace{-10pt}
% \vskip -0.2 in
\label{fig:figure1}
\end{figure}

However, existing VLMs for 3D medical imaging often struggle to consistently outperform vision-only models, failing to effectively leverage textual information to enhance segmentation performance~\cite{bassi2024touchstone}. This limited performance stems from the inability to effectively integrate vision and language features into a cohesive shared representation. Two key challenges drive this: (1) the inherent differences between vision and language representations, which make seamless unification difficult, and (2) the complexity and domain-specific nature of medical terminology, which off-the-shelf text encoders struggle to capture, ultimately weakening the contribution of textual features to segmentation accuracy.

VLMs unify multimodal representations through various fusion strategies to process visual and textual data for a single task. Existing architectures primarily adopt three approaches (\Cref{fig:figure1}a (i–iii)): (1) fusion at the vision decoder~\cite{zhong2023ariadne, zeng2024abp, guo2024common, zhang2024madapter}, (2) late fusion after both encoders complete feature extraction~\cite{oh2023llm,zhao2023one,liu2023clip,jiang2024zept,wang2022cris}, or (3) injecting text embeddings only in the latter stages of the vision encoder~\cite{feng2021encoder,yang2022lavt,oh2023llm,li2023lvit}. The first two approaches keep the vision encoder unimodal until fusion, weakening cross-modal alignment due to the absence of early interaction. The third approach misses enriched representations from earlier text encoder stages, failing to capture crucial cross-modal cues for effective alignment~\cite{cho2023cross}. Consequently, intermediate-level features—where vision embeddings transition from low- to high-level semantics~\cite{dosovitskiy2020image,hatamizadeh2021swin} and text embeddings evolve from structural to global concepts~\cite{tenney2019bert, peters2018dissecting}—remain underutilized. To address this, we propose bidirectional progressive fusion, an architectural innovation enabling continuous information exchange at all encoder stages. By progressively integrating complementary features, our approach enhances multimodal representations, leading to a more cohesive cross-modal understanding (\Cref{fig:figure1}a(iv)).

However, fusion may be ineffective if the text encoder fails to generate meaningful embeddings, as off-the-shelf models often struggle with complex medical terminology, hindering alignment with vision features~\cite{huang2024cat, oh2023llm}. To address this, we propose a novel global-local alignment strategy that enhances cross-modal correspondence and improves text encoder comprehension, resulting in more meaningful embeddings. This extends vanilla contrastive learning~\cite{chen2020simple, liu2023clip} into a balanced coarse and fine-grained process, strengthening information exchange between encoders. As shown in the t-SNE plots in \Cref{fig:figure1}b, ClinicalBERT~\cite{alsentzer2019publicly} (left) can separate simple concepts like titles and locations, but struggles with domain-specific terms, whereas BiPVL-Seg (right) achieves improved concept separation through global-local alignment. Global alignment associates class-level descriptions with vision embeddings, while local alignment refines fine-grained concept associations with specific visual regions, ensuring both holistic class comprehension and precise concept differentiation. Existing VLM-based approaches that address this, typically rely on global alignment applied as a separate pre-training stage~\cite{zhao2023one,jiang2024zept,chen2023generative}, which decouples contrastive alignment from downstream segmentation and weakens the synergy between cross-modal alignment and segmentation performance.

Hence, we propose \textbf{BiPVL-Seg}, \textbf{Bi}directional \textbf{P}rogressive \textbf{V}ision-\textbf{L}anguage Fusion with global-local Alignment for medical image \textbf{Seg}mentation, an end-to-end framework that enhances any standard encoder-decoder vision model and text encoder to overcome the limitations of existing VLMs. Our contributions are three-fold: (1) we introduce a novel bidirectional progressive fusion architecture that enables progressive vision-language interaction within both encoders, (2) we propose a global-local contrastive alignment strategy directly integrated with segmentation training to improve multimodal representation learning, and (3) we demonstrate state-of-the-art performance across multiple medical segmentation benchmarks, covering diverse anatomical structures, tumors, and imaging modalities.

%% file: sec/2_related_work.tex
\section{Related Work}
\label{sec:related_work}

\subsection{Vision-Language Fusion Architectures}
While text integration into vision-only architectures has been explored in natural image segmentation~\cite{liang2023open,xu2022simple,luddecke2022image, li2024deal, cho2023cross}, its use in medical imaging, especially for 3D data, remains underexplored. Existing VLM architectures for vision-language fusion can be grouped into three main categories.

\noindent \textbf{Fusion at Vision Decoder} The simplest architecture for fusing vision-language information is to perform fusion at the decoder stage, where textual information assists in generating the final segmentation maps. Works such as~\cite{zeng2024abp, shastri2024locate, huemann2024contextual} follow this approach. In this category, the vision and text encoders process their modalities independently, with no interaction between them until the decoder. This limits the model’s ability to capture cross-modal dependencies during feature extraction, reducing the effectiveness of learned representations. Since feature extraction occurs in the encoder, early fusion is crucial, while delaying fusion to the decoder stage hinders effective cross-modal learning.

\noindent \textbf{Late Fusion at Vision Encoder}  
A slightly more refined architecture performs fusion within the vision encoder, but only at the final stages. Both encoders still process their modalities independently for most of the pipeline, limiting early cross-modal interaction. Simpler 3D VLMs~\cite{liu2023clip, li2024novel} using late fusion at the vision encoder rely on short text phrases, simplifying fusion to improve segmentation performance. \cite{liu2023clip} focuses on universal segmentation, while \cite{li2024novel} addresses mixed CT-MR training with text guidance. However, these methods do not address the challenge of processing more complex, domain-specific text inputs, which are far more common in real clinical scenarios.

To handle more complex scenarios, advanced 3D VLMs~\cite{zhao2023one, jiang2024zept, huang2024cat} enhance simple textual techniques by incorporating detailed class definitions. These definitions function more like fixed class profiles, where a static, predefined set of attributes describes each class. Despite adopting intricate fusion strategies, these methods rely on late fusion within the vision encoder, preventing intermediate-stage interaction. This leads to poorly aligned, disparate embeddings that hinder effective fusion.

\noindent \textbf{Unidirectional Fusion at Vision Encoder}
Other than late fusion, works such as~\cite{li2023lvit, hu2024lga, bui2024visual, zhong2023ariadne, guo2024common} in 2D and~\cite{oh2023llm, kim2023ro} in 3D adopt unidirectional fusion, where textual information is gradually injected into the vision encoder. These works incorporate patient-specific, clinician-generated textual information instead of generalized class definitions. However, fusion occurs only within the vision encoder, leaving the text encoder independent and unrefined. As a result, the vision encoder processes visual features enriched with textual information, while the text encoder lacks complementary visual feedback, preventing mutual refinement. Additionally, these methods rely solely on later-stage text embeddings, neglecting intermediate text encoder processing, further limiting the formation of a coherent multimodal representation. Unlike these approaches, our method enables stronger cross-modal interaction through bidirectional progressive fusion, enriching each encoder with complementary information. This ensures both operate within a unified space from the outset.

\subsection{Vision-Language Embeddings Alignment}
To enhance a vision task, auxiliary text embeddings must be meaningful enough to directly improve performance. Most of the above-mentioned works rely on domain-specific text encoders~\cite{lee2020biobert, alsentzer2019publicly, gu2021domain}, pre-trained for medical domains, as off-the-shelf components. However, works such as~\cite{chen2023generative, jiang2024zept, ding2024cat, zhao2023one} further introduce contrastive alignment to make text encoders more adaptive to specific tasks and datasets, enabling them to generate more meaningful multimodal embeddings. These approaches align a single, fixed sentence describing each organ or class with its vision embeddings using contrastive learning (referred to as global alignment), but this can lead to overly rigid alignment. This rigidity risks the text encoder memorizing specific sentence patterns rather than generalizing to shared, meaningful features.

We extend contrastive learning to a global-local strategy, aligning full textual descriptions with visual features globally and fine-grained concepts with corresponding regions locally. While fine-grained concept alignment has been explored in natural images~\cite{gao2022pyramidclip, li2024deal}, it remains largely unexplored in medical imaging. Unlike existing pipelines that rely on separate pretraining for modality alignment followed by finetuning for segmentation, we propose a unified training strategy that balances both within a single end-to-end process, ensuring joint optimization of multimodal alignment and segmentation.

%% file: sec/3_method.tex
\section{Method}

\begin{figure*}[t]
\centering
\includegraphics[width=0.85\textwidth]{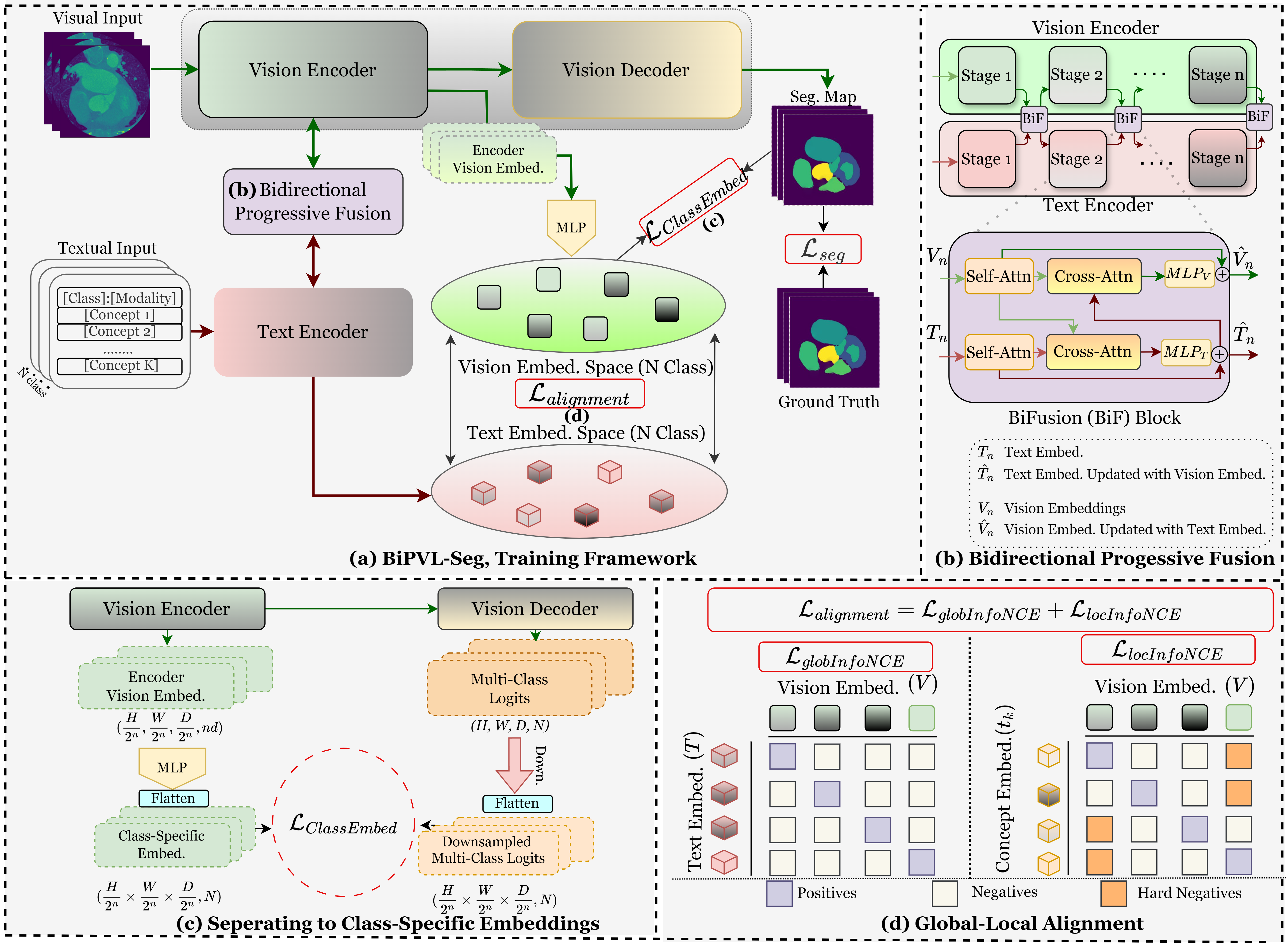}
% \vspace{-10pt}
\caption{(a) \textbf{BiPVL-Seg}, an end-to-end training pipeline with segmentation, class embedding, and global-local alignment losses. (b) \textbf{Bidirectional progressive fusion} between encoders via BiFusion blocks at each stage. (c) $\mathcal{L}_{ClassEmbed}$: Mapping encoder embeddings to class-specific embeddings with decoder supervision. (d) $\mathcal{L}_{alignment}$: Balanced global-local alignment linking class-level text embeddings to visual features, while fine-grained concept-level alignment uses hard negatives.}
\label{fig:figure2}
% \vskip -0.2 in
\end{figure*}

\begin{figure}[t]
\centering
\includegraphics[width=0.9\columnwidth]{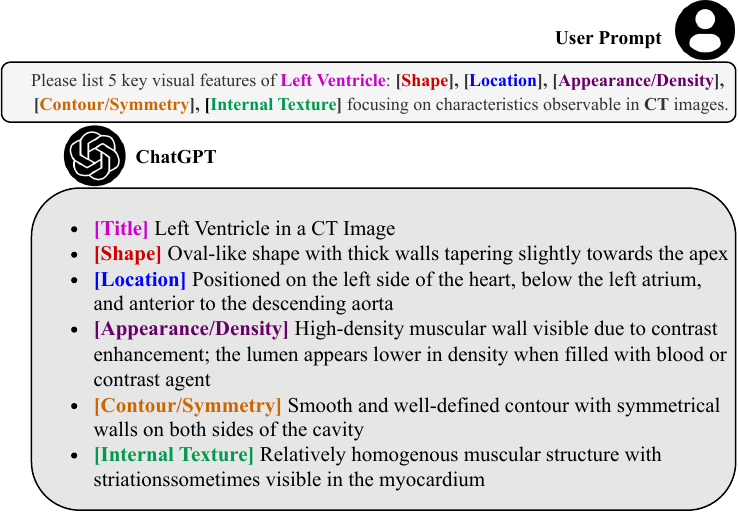}
\vspace{-10pt}
\caption{A demo of curating concept descriptions of each class in our datasets using ChatGPT.}
\vspace{-20pt}
% \vskip -0.2 in
\label{fig:figure3}
\end{figure}

We describe BiPVL-Seg, a new VLM segmentation framework that integrates vision-language representations through BiFusion, a bidirectional progressive fusion in the model architecture. It is further reinforced with a novel global-local embedding alignment at both class and concept levels within an end-to-end training pipeline.

\label{sec:method}
\subsection{Vision-Language Information Processing}
\label{language}
% Need to recheck the notations for uniformity

\noindent\textbf{Textual Input Generation} To serve as the textual input, semantically rich textual descriptions are generated by defining each class using visual concepts i.e. attributes of the class, following a consistent and fixed structure across datasets. The “Chat Completions API” of GPT-4o~\cite{openai2023gpt} is used to generate a predefined set of visually observable fine-grained concepts for each class. Each description follows the template: “[class], [shape], [location], [appearance/density], [contour/symmetry], [internal texture] in [modality]”. The LLM is explicitly instructed to focus on visually descriptive language, avoiding functional or diagnostic attributes, ensuring both modalities operate on aligned visual features. All generated concepts undergo manual review to ensure accuracy, clarity, and precision. \Cref{fig:figure3} illustrates a simplified prompt, with the complete set of prompts and class definitions provided in the Github Repository.

\noindent\textbf{Text Encoder}  Transformer-based language encoders~\cite{devlin2018bert, vaswani2017attention} process text holistically without distinct spatial stages, unlike vision encoders that naturally define stages through resolution changes. This mismatch complicates direct fusion between visual and text features at multiple stages.  To address this, we design a stage-divided BERT~\cite{devlin2018bert} encoder, structured to mirror the hierarchical stages of the vision encoder. Based on empirical evidence, we follow~\cite{cho2023cross} and partition the layers into four stages: [1–6] as stage 1, [7–8] as stage 2, [9–10] as stage 3, and [11–12] as stage 4. Each successive stage captures increasingly abstract and contextualized representations~\cite{tenney2019bert}, aligning with the progressive abstraction in the vision encoder. This design enables progressive multi-stage fusion between modalities, matching representations across corresponding encoder depths. Our text encoder can be initialized with any domain-specific BERT checkpoint~\cite{lee2020biobert, alsentzer2019publicly, gu2021domain}, allowing flexibility across medical and general-purpose applications. The text encoder consists of 12 transformer layers, an additional input embedding layer, and operates with a feature dimension of 768.

\noindent\textbf{Vision Model}
A standard encoder-decoder 3D model like Swin UNETR~\cite{he2023swinunetr} is used, where the vision encoder extracts embeddings progressively (\Cref{fig:figure2}a). It consists of four stages with Swin Transformer blocks and downsampling layers. Extracted embeddings, enriched with text, are passed to the decoder via skip connections and concatenated with encoder outputs. The main segmentation loss ($\mathcal{L}_{\text{seg}}$) from \Cref{fig:figure2}a  can be defined as a uniform combination of Dice ($\mathcal{L}_{\text{dice}}$) and Cross-Entropy loss ($\mathcal{L}_{\text{ce}}$):
\begin{equation}
    \mathcal{L}_{\text{seg}} = \mathcal{L}_{\text{dice}}(P, G) + \mathcal{L}_{\text{ce}}(P, G), 
\label{equ:task1}
\end{equation}

\noindent where $P$ represents the predicted segmentation map derived from the decoder's logits, and \(G\) denotes the corresponding ground truth mask.

\subsection{BiFusion: Bidirectional Progressive Fusion}
The BiFusion block at each encoder stage enables bidirectional progressive fusion by applying cross-modal attention, allowing each modality to refine its features while integrating information from the other. As visual and textual inputs traverse their respective encoders, they undergo fusion at each stage $n$ ($n \in \{1,2,3,4\}$), ensuring continuous cross-modal interaction. Let $\mathbf{T}_n$ and $\mathbf{V}_n$ denote the text and vision embeddings at stage $n$, with shapes $(N, 768)$ and $\left(\frac{H}{2^n},\frac{W}{2^n}, \frac{D}{2^n}, nd\right)$, respectively, where $d$ represents the feature dimension of the vision model.  The BiFusion block first flattens $\mathbf{V}_n$ to shape $\left(\frac{H}{2^n}\times\frac{W}{2^n}\times\frac{D}{2^n}, nd\right)$ for compatibility with the text embeddings. Since visual features are progressively downsampled across stages, while text embeddings remain at a fixed 768 dimensions in the BERT encoder, a learnable linear layer maps text embeddings to match the visual feature dimensions at each stage before the fusion. 

The progressive fusion, shown in \Cref{fig:figure2}b, begins with a LayerNorm (LN) followed by a Self-Attention (SA) operation applied independently to both modalities, producing $\mathbf{T'}_n$ and $\mathbf{V'}_n$. This step captures intra-modal contextual relationships within each modality.  Next, bidirectional Cross-Attention (CA) is applied to exchange information across modalities. In the first CA step, $\mathbf{T'}_n$ serves as queries, while $\mathbf{V'}_n$ acts as keys and values, enabling the text embeddings to attend to spatially relevant visual features. An MLP layer refines the result, producing updated text embeddings $\mathbf{\hat{T}}_n$ enriched with visual information. In the second CA step, roles are reversed: $\mathbf{V'}_n$ serves as queries, while $\mathbf{\hat{T}}_n$ acts as keys and values, producing updated vision embeddings $\mathbf{\hat{V}}_n$ that integrate complementary text-derived context. 

\vspace{-2mm}
\begin{equation}
\begin{aligned}
\mathbf{T'_n} &= \text{SA}(LN(\mathbf{T_n})), \mathbf{V'_n}=\text{SA}(LN(\mathbf{V_n})),\\
\mathbf{\hat{T}}_{n} &= \mathbf{T'_n}+\text{MLP}_T(\text{CA}(\mathbf{T'}_{n}, \mathbf{V'}_n)),\\
\mathbf{\hat{V}}_n &= \mathbf{V'_n}+\text{MLP}_V(\text{CA}(\mathbf{V'}_n, \mathbf{\hat{T}}_{n})). 
\end{aligned}
\end{equation}

\noindent The updated embeddings $\mathbf{\hat{T}}_n$ and $\mathbf{\hat{V}}_n$ are passed to stage $n+1$ of their respective encoders, ensuring progressive cross-modal fusion until the final stage.

\subsection{Vision-Language Global-Local Alignment} 
To enhance cross-modal representation learning, we introduce a global-local alignment strategy that enables both encoders to generate semantically aligned embeddings in a shared space. This strategy comprises coarse global alignment, linking high-level textual class descriptions to global visual representations, and fine-grained local alignment, associating individual visual features with detailed textual concepts.

\subsubsection{Class-Specific vision Embedding Supervision}

During training, we enforce alignment between class-specific vision embeddings from the final stage of the vision encoder and the corresponding class-specific text embeddings from the final stage of the text encoder. Initially, vision embeddings extracted from the encoder have the shape $\left(\frac{H}{2^n}, \frac{W}{2^n}, \frac{D}{2^n}, nd\right)$. 
To obtain class-specific representations, we apply an MLP block at vision encoder’s output, which maps these generalized embeddings $\left(\frac{H}{2^n}, \frac{W}{2^n}, \frac{D}{2^n}, nd\right)$ into class-specific representations with shape $\left(\frac{H}{2^n}, \frac{W}{2^n}, \frac{D}{2^n}, N\right)$, where each channel corresponds to a specific class embedding. This transformation is guided by $\mathcal{L}_{\text{ClassEmbed}}$, a Cross-Entropy loss that supervises the mapping process using class-specific logits obtained from the vision decoder (illustrated in \Cref{fig:figure2}c). 

By using the decoder’s logits to supervise the encoder-side embeddings, the model progressively refines visual features, ensuring they correctly map into class space throughout training. Let $P$ denote the class-specific logits from the decoder, containing $N$ class embeddings. These logits are downsampled to match the spatial resolution of the generalized embeddings. Both the vision embeddings and the class logits are then flattened before computing the Cross-Entropy loss, ensuring that the mapping from $nd$ channels to $N$ class-specific channels is directly supervised by the decoder’s predictions. As training progresses, the refined logits from \Cref{equ:task1} further enhance class-specific consistency within the encoder.

\vspace{-4mm}
\begin{equation}
    \mathcal{L}_{\text{ClassEmbed}} = \text{CE}(\text{MLP}(\mathbf{V}_n), P)
    \label{equ:task2}
\end{equation}
\vspace{-5mm}

\subsubsection{Coarse Global Alignment}
For coarse global alignment, aggregated text embeddings ($\mathbf{T}$) and vision embeddings ($\mathbf{V}$) are aligned for each of the \(N\) classes using a contrastive loss. As detailed in \Cref{language}, each class is associated with a fixed set of concept descriptions. To map these descriptions into class space, we extract concept-level embeddings (\(\mathbf{t}_k\), where \(k \in [1, K]\) and \(K\) is the total number of concepts) using the text encoder. Both \(\mathbf{t}_k\) and $\mathbf{V}$ are projected into a common feature space via learnable linear projection layers, followed by $L2$ normalization. To form each class-level text embedding ($\mathbf{T}$), we compute learnable weights \(w_k\) that emphasize concept embeddings most relevant to their visual counterpart. These weights are derived from the cosine similarity between each projected concept embedding ($\mathbf{t}'_k$) and its corresponding class vision embedding ($\mathbf{V'}$):

\vspace{-5mm}
\begin{equation}
\begin{aligned}
    \mathbf{t'}_k &= \text{norm}\{\text{Proj}_\mathbf{t}(\mathbf{t}_k)\}, \quad
    \mathbf{V'} = \text{norm}\{\text{Proj}_\mathbf{v}(\mathbf{V})\} \\ 
    w_k &= \text{softmax}\{\text{cos}(\mathbf{t'}_k, \mathbf{V'})\}, \mathbf{T} = \sum_{k=1}^{K} w_k \mathbf{t}_k.
\end{aligned}
\end{equation}
\vspace{-5mm}

\noindent The global alignment loss, $\mathcal{L}_{\text{globInfoNCE}}$, is then computed using an InfoNCE loss, which enforces bidirectional alignment between each class pair of text embeddings $\mathbf{T}_j$ and vision embeddings $\mathbf{V}_i$, for all classes $i, j \in [1, N]$. To prevent premature convergence and promote more robust alignment, we apply small Gaussian noise perturbations to both $\mathbf{T}_j$ and $\mathbf{V}_i$, ensuring the embeddings do not become overconfident early in training. We compute bidirectional similarity matrices: vision-to-text ($V \rightarrow T$) and text-to-vision ($T \rightarrow V$), using normalized dot product similarity:

\vspace{-5mm}
\begin{equation}
    \mathbf{S}_{ij}^{V \rightarrow T} = \frac{\mathbf{V}_i \cdot \mathbf{T}_j}{\|\mathbf{V}_i\| \|\mathbf{T}_j\|}, \quad \mathbf{S}_{ij}^{T \rightarrow V} = \frac{\mathbf{T}_i \cdot \mathbf{V}_j}{\|\mathbf{T}_i\| \|\mathbf{V}_j\|}.
    \label{equ:similarity}
\end{equation}
\vspace{-3mm}

\noindent , where $\tau$ denotes the temperature scaling parameter:

\begin{equation}
\begin{aligned}
    \mathcal{L}_{\text{InfoNCE}^{V \rightarrow T}} &= -\frac{1}{N} \sum_{i=1}^{N} \log \frac{\exp(\mathbf{S}_{ii}^{V \rightarrow T}/\tau)}{\sum_{j=1}^{N} \exp(\mathbf{S}_{ij}^{V \rightarrow T}/\tau)}, \\
    \mathcal{L}_{\text{InfoNCE}^{T \rightarrow V}} &= -\frac{1}{N} \sum_{i=1}^{N} \log \frac{\exp(\mathbf{S}_{ii}^{T \rightarrow V}/\tau)}{\sum_{j=1}^{N} \exp(\mathbf{S}_{ij}^{T \rightarrow V}/\tau)}, \\
    \label{equ:alignment}
    \mathcal{L}_{\text{globInfoNCE}} &= \frac{1}{2}(\mathcal{L}_{\text{InfoNCE}^{V \rightarrow T}} + \mathcal{L}_{\text{InfoNCE}^{T \rightarrow V}}).
\end{aligned}
\end{equation}

\noindent The global InfoNCE loss, $\mathcal{L}_{\text{globInfoNCE}}$, maximizes the similarity of positive pairs $\mathbf{S}_{ii}$ (diagonal entries) while minimizing the similarity of negative pairs $\mathbf{S}_{ij}$ (off-diagonal entries) within the similarity matrix shown in \Cref{fig:figure2}d. Positive pairs correspond to matching class-level vision and text embeddings, while negative pairs represent mismatched class embeddings.

\begin{figure*}[t]
\centering
\includegraphics[width=0.80\textwidth]{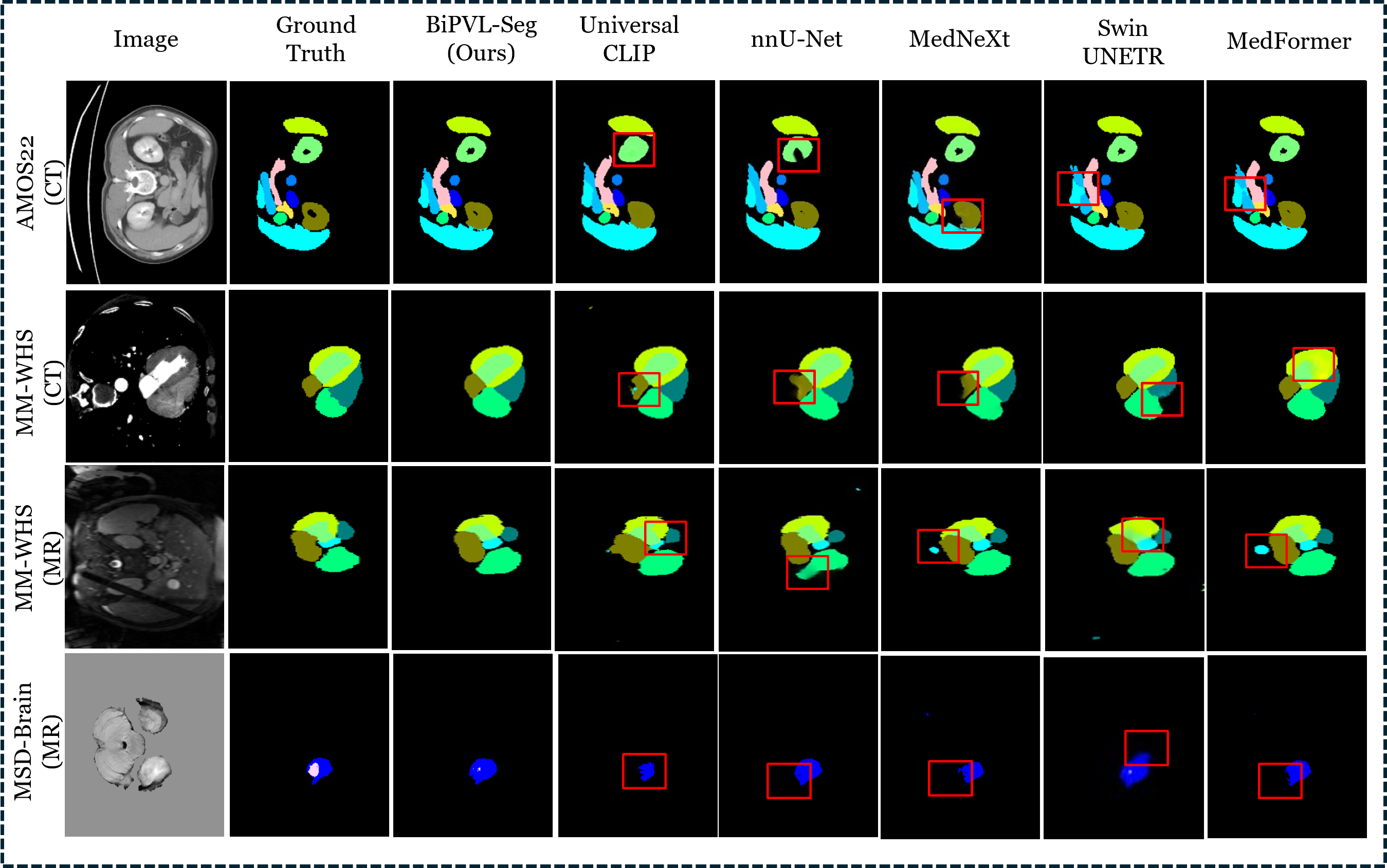}
% \vspace{-10pt}
\caption{Qualitative visualizations of randomly selected image slices from AMOS22, MM-WHS (CT), and MM-WHS (MR) and MSD-Brain (MR). Comparison of BiPVL-Seg against other models, with \textcolor{red}{red boxes} indicating areas where BiPVL-Seg outperforms.}
\label{fig:figure4}
% \vskip -0.2 in
\end{figure*}

\subsubsection{Fine-Grained Local Alignment}
To refine the text encoder’s understanding, we align individual textual concepts with their visual counterparts. While global alignment establishes coarse class-level associations, local alignment refines concept-level embeddings, preserving fine-grained semantics and preventing overfitting to fixed class labels. To perform fine-grained local alignment, we extend the bidirectional alignment framework from \Cref{equ:alignment} to operate on individual concept embeddings ($\mathbf{t}k$). Each concept embedding is aligned in both vision-to-concept ($V \rightarrow t_k$) and concept-to-vision ($t_k \rightarrow V$) directions, ensuring consistent positioning within the shared space. A key challenge in local alignment is the potential for incorrect supervision due to randomly sampled negatives, as multiple classes may share similar properties (e.g., shape, location). To address this, we select hard negatives from the top $\mathcal{K}$ most similar but incorrect concepts, ensuring meaningful contrastive learning (illustrated in \Cref{fig:figure2}d). For each concept-to-vision similarity matrix $\mathbf{S}{ij}^{t_k \rightarrow V}$, we identify the $\mathcal{K}$ most challenging negative pairs:

\vspace{-3mm}
\begin{equation}
\begin{aligned}
    \mathcal{N}_\mathcal{K} &= \underset{j}{\text{arg top-}}\mathcal{K} \left( \mathbf{S}_{ij}^{t_k \rightarrow V} \right), \quad \forall i \in \{1, \dots, N\}, \\
    \mathcal{L}_{\text{InfoNCE}^{t_k \rightarrow V}} &= -\frac{1}{N} \sum_{i=1}^{N} \log \frac{\exp(\mathbf{S}_{ii}^{t_k \rightarrow V}/\tau)}{\sum_{j \in \mathcal{N}_\mathcal{K}} \exp(\mathbf{S}_{ij}^{t_k \rightarrow V}/\tau)}.
\end{aligned}
\end{equation}

\noindent A similar process is applied to compute $\mathcal{L}_{\text{InfoNCE}^{V \rightarrow t_k}}$, where visual features attend to concept-level embeddings. The final fine-grained local alignment loss $\mathcal{L}_{\text{locInfoNCE}}$ averages these bidirectional losses across all $K$ concepts:
\vspace{-3mm}
\begin{equation}
    \mathcal{L}_{\text{locInfoNCE}} = \frac{1}{2K} \sum_{k=1}^{K} \left( \mathcal{L}_{\text{InfoNCE}^{V \rightarrow t_k}} + \mathcal{L}_{\text{InfoNCE}^{t_k \rightarrow V}} \right).
\end{equation}
% \vspace{-3mm}
\noindent The final alignment loss $\mathcal{L}_{\text{alignment}}$ integrates both coarse-grained and fine-grained alignment terms, balanced by weighting parameters $\alpha_1$ and $\alpha_2$, which controls the relative contribution of global and local alignment:
\vspace{-2mm}
\begin{equation}
     \mathcal{L}_{\text{alignment}} = \alpha_1 \mathcal{L}_{\text{globInfoNCE}} + \alpha_2 \mathcal{L}_{\text{locInfoNCE}}.
     \label{equ:task3}
\end{equation}

\subsection{End-to-End Pipeline} 
The three main tasks of the pipeline from \Cref{equ:task1}, \Cref{equ:task2}, and \Cref{equ:task3} are optimized within a single training loop, enabling end-to-end efficiency, as illustrated in \Cref{fig:figure2}a. The overall loss function is formulated as:

\vspace{-7mm}
\begin{equation}
    \mathcal{L} = \beta_1\mathcal{L}_{\text{seg}}+\beta_2\mathcal{L}_{\text{ClassEmbed}}+\beta_3\mathcal{L}_{\text{alignment}},
\end{equation}

\noindent where $\beta_1$, $\beta_2$, and $\beta_3$ are learnable parameters, adjusted dynamically through homoscedastic uncertainty~\cite{kendall2018multi}, ensuring task-dependent weighting for optimal multi-task learning.

%% file: sec/4_experiments.tex
\section{Experiments}

\begin{table*}[t]
\centering

% \scriptsize
\small
\resizebox{\textwidth}{!}{
\begin{tabular}{l|ccc|ccc|ccc|ccc}
\toprule
        % \hline
        \multirow{3}{*}{Method}&\multicolumn{3}{c|}{AMOS22 (CT)}&\multicolumn{3}{c|}{MM-WHS (CT)}&\multicolumn{3}{c|}{MM-WHS (MR)}&\multicolumn{3}{c}{MSD-Brain (MR)}\\
        \cline{2-13}
        &DSC$\uparrow$&HD95$\downarrow$&NSD$\uparrow$&DSC$\uparrow$&HD95$\downarrow$&NSD$\uparrow$&DSC$\uparrow$&HD95$\downarrow$&NSD$\uparrow$&DSC$\uparrow$&HD95$\downarrow$&NSD$\uparrow$\\
        % \cline{2-4} \cline{5-7} \cline{8-10} \cline{11-12}
        \hline
        \hline 
         U-Net~\cite{ronneberger2015unet} & 80.64 & 8.88 & 87.99 & 89.19&3.94&93.62&76.33&\underline{28.44}&76.61&68.73 & 15.70 & 73.33\\
        % \hline
        UNet++~\cite{zhou2018unet++}& 84.30 & 11.46 & 91.14 & 90.29&4.14&93.91&79.37&42.85&78.42& 74.10 & 12.05 & 76.98 \\
        % ATTUNet~\cite{oktay2018attention} & 69.83 & 12.51 & 74.56 & 89.93 & 3.87 & 93.68&75.37&39.66&71.81& - & - & -\\
        % \hline
        % \hline
        nnU-Net~\cite{isensee2021nnu}& \underline{86.33} & 9.05 & 92.01 & 90.05&4.03&93.72&\underline{80.71}&46.37&79.21& 75.10 & 12.38 & 78.29\\
        STU-Net~\cite{huang2023stu}& 84.21 & 8.88 & 91.42 & 90.85&3.88&94.22&77.11&37.51&77.25&72.05&11.95&77.97\\
        MedNeXt~\cite{roy2023mednext}& 84.88 & 8.32 & 91.92 & 90.39&3.62&94.11&77.54&36.51&77.11&71.84&14.59&76.90\\
        \hline
        UNETR ~\cite{hatamizadeh2022unetr} & 77.67 & 12.20 & 83.22 & 88.96&4.09&91.84&74.72&30.46&75.44& 70.78 & 14.25 & 75.11\\
        % \hline
        Swin UNETR ~\cite{hatamizadeh2021swin} & 84.71 & 10.84 & 90.74 & 90.48&3.83&93.83&80.46&37.37&80.40&\textbf{75.82} & 12.31 & 78.76\\
        SwinUNETR-V2 ~\cite{he2023swinunetr} & 85.53 & 9.28 & 91.75
        & 90.51&3.85&94.34&80.04&35.52&\underline{81.08}&75.02 & 12.70 & 77.63\\
        nnFormer~\cite{zhou2021nnformer}& 76.98 & 10.98 & 84.57 & \underline{91.43}&\underline{3.52}&\underline{95.11}&74.54&36.75&77.03&72.44&12.83&77.55\\
        MedFormer~\cite{gao2022data}& 85.84 & 8.83 & \underline{92.03} & 90.68&3.79&93.83&78.18&37.88&78.64&74.64&\underline{10.32}&78.20\\
        
        \hline
        \hline
        Universal~\cite{liu2023clip} & 82.59 & 11.25 & 88.02 & 90.14&4.23&93.27&77.07&44.30&76.71&72.31&12.65&75.85\\
        % \hline
        % \hline
        % \hline
        MulModSeg~\cite{li2024novel}& 83.79 & \underline{8.28} & 89.56 & 89.95&4.08&93.41&80.23&34.54&80.46&72.65&12.24&76.54\\
        ZePT~\cite{jiang2024zept}& 81.12 & 13.02 & 85.35 & 89.96&4.94&93.17&77.56&33.01&78.25&73.47&14.88&\underline{79.28}\\
        CAT~\cite{huang2024cat}& 83.35 & 9.56 & 88.55 & 90.22&3.98&93.11&77.01&38.17&79.25&71.14&15.64&77.19\\
        BiPVL-Seg (Ours)  & \textbf{88.21} & \textbf{6.52} & \textbf{93.97} & \textbf{92.60}&\textbf{3.09}&\textbf{96.86}&\textbf{82.94}&\textbf{27.01}&\textbf{83.14}&\underline{75.42}&\textbf{9.82}&\textbf{79.87}\\
        % \hline
        % \hline
        \bottomrule
    \end{tabular}
    }
    \caption{Comparison of BiPVL-Seg with benchmark vision-only models and VLMs for medical image segmentation. Vision-only models appear first, followed by VLMs (separated by a double horizontal line). The best results are \textbf{bolded} and the second-best results are \underline{underlined}.} 
    
    \vspace{-2mm}
    \label{tab:results}
\end{table*}

Our objective is to validate the effectiveness of the newly proposed BiPVL-Seg, a multimodal framework designed to enhance medical image segmentation through various metrics. This will be achieved by conducting a comprehensive set of experiments aimed at addressing key research questions: \textbf{\textit{Q1}}: Does BiPVL-Seg outperform the existing benchmark vision-only models? \textbf{\textit{Q2}}: How does BiPVL-Seg perform compared to the existing benchmark VLMs? \textbf{\textit{Q3}}: How does the inclusion of textual information impact overall performance?

\label{sec:experiments}

\subsection{Implementation Details}
\label{evaluation}
\noindent\textbf{Datasets}
We selected 3D imaging datasets spanning CT and MRI for diverse modalities and anatomical structures: AMOS22 (CT)~\cite{ji2022amos} (abdominal organs), MM-WHS (CT/MR)~\cite{zhuang2019evaluation} (cardiac substructures), and MSD-Brain (MR)~\cite{antonelli2022medical} (brain tumors). Additional dataset and preprocessing details are available in the appendix.

\noindent \textbf{Experiments Setup} For our experiments, we utilized Hugging Face's Transformers library~\cite{wolf2019huggingface} to initialize our text encoder with ClinicalBERT~\cite{alsentzer2019publicly}'s weights and Swin UNETR~\cite{he2023swinunetr} as the vision model. Each model was trained over $300$ epochs with validation performed every $5$ epochs. We used the AdamW optimizer with a learning rate of $10^{-4}$ and weight decay of $10^{-5}$. We applied a cosine annealing scheduler to adapt the learning rate dynamically, which decayed the maximum learning rate smoothly to a minimum value of $10^{-7}$ over the training period. More details on the hyperparameters tuning are in the appendix.  All the experiments were conducted on an NVIDIA GeForce RTX 4090 GPU with 24 GB of memory using Python $3.9.18$.

\noindent \textbf{Benchmark Models:} We select leading semantic segmentation models for medical imaging, including U-Net-style (CNN-based) models (U-Net~\cite{ronneberger2015unet}, UNet++~\cite{zhou2018unet++}, nnU-Net~\cite{isensee2021nnu}, STU-Net~\cite{huang2023stu}, and MedNeXt~\cite{roy2023mednext}) and UNETR-style (Vision Transformer-based) models (UNETR~\cite{hatamizadeh2022unetr}, Swin UNETR~\cite{hatamizadeh2021swin}, SwinUNETR-V2 ~\cite{he2023swinunetr}), nnFormer~\cite{zhou2021nnformer}, and MedFormer~\cite{gao2022data}). To further compare BiPVL-Seg with popular VLMs, we pick the VLMs that are closest to our work such as the Clip-driven Universal Model~\cite{liu2023clip}, MulModSeg~\cite{li2024novel}, ZePT~\cite{jiang2024zept}, and CAT~\cite{huang2024cat}. All models were trained from scratch with three-fold cross-validation using default settings. 

\noindent \textbf{Evaluation Metrics} We assess model performance using the Dice Similarity Coefficient (DSC, \%), 95\% percentile Hausdorff Distance (HD95, mm), and Normalized Surface Distance (NSD, \%)~\cite{nikolov2018deep}.

\subsection{Results and Discussion}

\noindent\textbf{BiPVL-Seg's Superiority Over Vision-Only Methods (\textbf{\textit{Q1}}):}
\Cref{fig:figure4} shows qualitative comparisons between BiPVL-Seg, top-performing vision-only models, and VLMs. Each row presents randomly selected image slices, ground truth labels, and segmentation outputs from these models, including BiPVL-Seg. The first row shows results from AMOS22 (CT), the second from MSD-Brain (MR), and the last two rows from MM-WHS (CT) and MM-WHS (MR), respectively. As highlighted in \Cref{fig:figure4}, BiPVL-Seg consistently produces outputs closer to the ground truth, with red boxes marking challenging regions where BiPVL-Seg outperforms both vision-only and VLM baselines, particularly in low-contrast areas and regions with ambiguous boundaries. We attribute this improvement to BiPVL-Seg’s ability to capture fine-grained concepts through global-local alignment, combined with the complementary information introduced via textual features.

We refer to \Cref{tab:results} for quantitative comparisons between BiPVL-Seg and top-performing vision-only models. BiPVL-Seg consistently outperforms all vision-only baselines across multiple datasets, demonstrating superior boundary refinement and segmentation accuracy. We attribute this to the additional meaningful text embeddings that directly enhance segmentation performance. BiPVL-Seg excels on AMOS22, outperforming the second-best model by nearly 2\% in DSC, 2 mm in HD95, and 2\% in NSD. It also achieves the best HD95 and NSD on MSD-Brain and outperforms all models across metrics in both MM-WHS datasets. Its strong performance across CT and MR, as well as in organ, heart substructure, and tumor segmentation, underscores its generalizability in medical imaging.

\noindent\textbf{BiPVL-Seg's Comparison with Other VLMs (\textbf{\textit{Q2}}):}
We also compare BiPVL-Seg with popular established VLMs in \Cref{tab:results}, where BiPVL-Seg outperforms all existing VLMs across datasets and modalities, including a notable 4\% improvement in both DSC and NSD over the second-best VLM on AMOS22. Notably, none of the other VLM baselines consistently surpass vision-only models, partly due to task-specific designs. For example, ZePT~\cite{jiang2024zept} focuses on unseen tumor segmentation, while MulModSeg~\cite{li2024novel} targets multi-domain mixed training across CT and MR. These specialized designs limit their performance on datasets with detailed anatomical annotations. In contrast, BiPVL-Seg serves as a general-purpose framework, excelling in multi-organ, heart substructure, and tumor segmentation across both CT and MR, establishing a strong benchmark for future multimodal segmentation research.

\begin{table}[t]
\centering

%\scriptsize
\small
\resizebox{1.0\columnwidth}{!}{
    \begin{tabular}{ccc|c|c|c}
        \toprule
        \multicolumn{3}{c|}{\multirow{2}{*}{Textual Information Format}}& \multirow{2}{*}{Alignment}&\multicolumn{2}{c}{AMOS22 (CT)}\\
        \cline{5-6}
        &&&&DSC$\uparrow$&NSD$\uparrow$\\
        \hline \hline
        \multicolumn{3}{c|}{Fixed, consistent concepts per class}  &Global-local&\textbf{88.23}&\textbf{93.97}\\
        \multicolumn{3}{c|}{Non-fixed, ambiguous concepts per class}   &Global-local& 87.24& 93.22\\
        \multicolumn{3}{c|}{Single aggregated definition per class}  &Global& 87.01& 92.91\\
        \multicolumn{3}{c|}{Class names only, no concepts}   &Global & 86.45& 91.34 \\
        \bottomrule
        
    \end{tabular}
    }

    \caption{Experimenting on the impact of different textual information formats in BiPVL-Seg. The best results are \textbf{bolded}.}
    \label{tab:ablation1}
    \vspace{-5mm}
\end{table}

\noindent\textbf{Effects of Textual Information in BiPVL-Seg (\textbf{\textit{Q3}}):}  
Since language enhances segmentation, we evaluate its impact on BiPVL-Seg using the AMOS22 (CT) dataset in \Cref{tab:ablation1}. We tested multiple formats of textual information as inputs: fixed consistent concepts (as described in \Cref{language}), non-fixed/ambiguous concepts (a variable set of unrefined concepts that have not undergone manual review to resolve ambiguity), a single detailed sentence combining all concepts into one description, and only class names as textual information. Examples of each format are provided in the appendix. Textual information with consistent, detailed concepts (row 1) achieves the best performance, while ambiguous concepts (row 2) lead to a clear drop. A single long definition per class (row 3) reduces DSC by nearly 2\%, suggesting potential memorization. Using only class names (row 4) results in a sharper 3\% drop, underscoring the need for detailed descriptions. Furthermore, \Cref{tab:ablation2} shows that removing all text inputs (leaving only the vision backbone) causes a nearly 4\% DSC drop, highlighting the importance of fixed concepts and BiPVL-Seg’s ability to effectively utilize them. 

\begin{table}[t]
\centering

%\scriptsize
\small
\resizebox{\columnwidth}{!}{
    \begin{tabular}{cccc|cc}
        \toprule
        \multicolumn{4}{c|}{BiPVL-Seg Components}& \multicolumn{2}{c}{AMOS22 (CT)}\\
        \hline
        \multirow{2}{*}{Backbone}&\multirow{2}{*}{BiFusion}&Global&Local &\multirow{2}{*}{DSC$\uparrow$}&\multirow{2}{*}{NSD$\uparrow$}\\
        &&Alignment&Alignment&&\\
        \hline \hline
        UNet++  & $\checkmark$ & $\checkmark$ & $\checkmark$&86.47&92.52\\
        Swin UNETR & $\checkmark$& $\checkmark$ & $\checkmark$ & \textbf{88.21}& \textbf{93.97}\\
        \hline
        Swin UNETR   & $\checkmark$& $\checkmark$ & $\times$& 87.54& 92.89\\
        Swin UNETR  & $\checkmark$& $\times$ & $\times$ & 86.71& 92.07 \\
        % Swin UNETR  & $\checkmark$& $\times$ & $\checkmark$ & 85.02 & 91.29 \\
        Swin UNETR   & $\times$& $\checkmark$ & $\checkmark$& 85.56& 91.66\\
        Swin UNETR  & $\times$& $\times$ & $\times$ & 84.71& 90.74 \\
        % $\times$  & $\times$ & $\times$   & -& -& -\\
        \bottomrule
    \end{tabular}
    }
    \caption{Ablation study: examining the effects of the performance based on various components in BiPVL-Seg. The best results are \textbf{bolded}.}
    \label{tab:ablation2}
    \vspace{-0.15in}
\end{table}

\noindent\textbf{Ablation Studies}  
We conduct ablation experiments to assess the contribution of each BiPVL-Seg component, as shown in \Cref{tab:ablation2}. First, we compare two vision backbones with all components applied, demonstrating significant improvement over their original results in \Cref{tab:results}. Additional experiments with other backbones are provided in the appendix. We then test: (1) replacing BiFusion with late fusion, (2) removing local alignment (retaining only global alignment, similar to traditional VLMs), (3) using off-the-shelf text encoder i.e. ClinicalBERT without any alignment, and (4) removing both BiFusion and alignment. Performance drops progressively, with DSC decreasing over 4\% and NSD over 2\% in the final configuration, highlighting the importance of both architectural and training innovations.

%% file: sec/5_conclusion.tex
\section{Conclusion} \label{sec:conclusion}
We present BiPVL-Seg, a vision-language framework for medical image segmentation that overcomes the limitations of existing VLMs. Current VLMs struggle to unify visual and textual information in a shared space, limiting their effectiveness. BiPVL-Seg addresses this with a bidirectional fusion architecture that progressively exchanges information between encoders, complemented by a global-local alignment strategy to enhance cross-modal embedding coherence.

\noindent\textbf{Limitations:} 
BiPVL-Seg has not been evaluated for universal segmentation of unseen organs, tumors, or imaging modalities, where generalization beyond annotated training distributions remains challenging. Extending it to handle such cases with improved robustness remains a key direction for future work.

% \newpage

%% file: sec/X_suppl.tex
\clearpage
\setcounter{page}{1}
\maketitlesupplementary

\section{Additional Technical Details and Benchmarks}
\label{sec:rationale}

\noindent\textbf{Vision Model}
We utilize any encoder-decoder-based 3D segmentation models, i.e. Swin UNETR~\cite{he2023swinunetr}, and utilize the vision encoder to initiate the stage-wise fusion. The vision encoder takes a 3D image, $X \in \mathbb{R}^{(H, W, D, 1)}$, where ($H, W, D$) represents the spatial resolution of gray-scale images and process the vision embeddings of the image sequentially. The encoder can be classified into $4$ stages, each containing multiple Swin Transformer~\cite{hatamizadeh2021swin} blocks with  MLPs, Layer Norm (LN), and skip connections, followed by a downsampler (except the final stage). At the end of each stage, downsampling of the vision embeddings occurs, and the resulting feature maps at each stage $n$ have sizes ($\frac{H}{2^n},\frac{W}{2^n},\frac{D}{2^n}$).

The encoder's extracted feature representations, which are gradually processed vision embeddings enriched with corresponding textual information, are utilized in the decoder via skip connections at each resolution. At each decoder stage, features are concatenated with outputs from corresponding encoder stages, facilitating the integration of both fine-grained details and high-level abstractions. In the final stage, the convolution layer is followed by a sigmoid activation function that maps outputs to probabilities, generating the final multi-class segmentation map.

\noindent \textbf{Text Encoder}  
We tested our framework with three different domain-specific BERT models:  

\begin{itemize}
    \item \textbf{BioBERT~\cite{lee2020biobert}}  
    A domain-specific language model pre-trained on biomedical text corpora using the BERT architecture, enabling enhanced contextual understanding of biomedical terms and relationships. By leveraging large-scale biomedical datasets such as PubMed abstracts and PMC full-text articles, BioBERT excels in tasks like named entity recognition, relation extraction, and question answering within the biomedical domain. Its ability to retain semantic nuances and adapt to downstream tasks makes it a foundational tool for integrating textual information into vision-language pipelines. This model’s robust representations of biomedical text empower cross-modal frameworks, improving alignment between textual and vision embeddings in medical imaging tasks.
    
    \item \textbf{ClinicalBERT~\cite{alsentzer2019publicly}}  
    ClinicalBERT is a language model specifically fine-tuned on electronic health records (EHRs) and clinical notes, building upon the BERT architecture to enhance understanding of medical terminology used in real-world clinical settings. Unlike BioBERT, which focuses on general biomedical literature, ClinicalBERT captures domain-specific patterns from structured and unstructured clinical text. This specialization enables improved performance in tasks such as clinical concept extraction, entity linking, and clinical decision support. Its ability to process nuanced clinical language makes it particularly useful for aligning textual data with medical imaging reports in vision-language applications.

    \item \textbf{PubMedBERT~\cite{gu2021domain}}  
    PubMedBERT is a domain-adaptive BERT model trained entirely from scratch on PubMed abstracts, offering deeper contextual representations of biomedical text. Unlike BioBERT, which starts from general BERT weights, PubMedBERT is pre-trained exclusively on biomedical data, allowing for improved generalization within the biomedical domain. This model demonstrates superior performance in biomedical NLP tasks such as text classification, information retrieval, and relation extraction. Its domain-specific embeddings make it a strong candidate for medical vision-language frameworks, aiding in more precise textual-visual alignment for downstream segmentation and classification tasks.
\end{itemize}

\noindent \textbf{Textual Information Prompt} 
To create the textual information, here is the full prompt we feed to GPT-4o (also, the whole set of prompts for each of the classes can be found alongside the code):

\begin{mdframed}[backgroundcolor=lightgray]
``[organ name] in a [modality] Image."

\noindent ``Shape: visual description of the shape of [organ name] in [modality] Image."

\noindent ``Location: visual description of the location of [organ name] in [modality] Image."

\noindent ``Appearance/Density: visual description of the appearance/density of [organ name] in [modality] Image."

\noindent ``Contour/Symmetry: visual description of the contour/symmetry: of [organ name] in [modality] Image."

\noindent ``Internal Texture: visual description of the internal texture of [organ name] in [modality] Image."

\noindent Can you provide six total descriptions for each class in this format?

\noindent Do it for the following organs, and the modality is CT: left ventricle, right ventricle, left atrium, right atrium, myocardium of LV, ascending aorta, pulmonary artery.

\noindent (We don't need functional descriptions, we need visual descriptions, please remember this throughout.) 
\end{mdframed}

\begin{table*}[ht]
\centering
\begin{tabular}{|l|c|c|c|}
\hline
\textbf{Dataset}         & \textbf{Voxel Spacing (mm)}    & \textbf{Intensity Range}       & \textbf{Normalization}                \\ \hline
\textbf{AMOS22 (CT)}     & \(1.5 \times 1.5 \times 2.0\)  & \([-175, 250]\) HU             & \([0, 1]\)                            \\ \hline
\textbf{MM-WHS (CT)}     & \(1.5 \times 1.5 \times 2.0\)  & \([0, 400]\) HU                & \([0, 1]\)                            \\ \hline
\textbf{MM-WHS (MR)} & \(1.5 \times 1.5 \times 2.0\) & Not applicable (MRI) & Non-zero voxel normalization, channel-wise                      \\ \hline
\textbf{MSD-Brain (MR)} & \(1.0 \times 1.0 \times 1.0\) & Not applicable (MRI) & Non-zero voxel normalization, channel-wise                      \\ \hline
                         
\end{tabular}
\caption{Dataset-Specific pre-processing details.}
\label{tab:dataset_processing}
\end{table*}

\noindent\textbf{The main segmentation loss $\mathcal{L}_{\text{seg}}$} is formulated as a uniform combination of the Dice loss \(\mathcal{L}_{\text{dice}}\) and the cross-entropy loss \(\mathcal{L}_{\text{ce}}\), given by:

\begin{equation}
\mathcal{L}_{\text{seg}} = \mathcal{L}_{\text{dice}}(P, G) + \mathcal{L}_{\text{ce}}(P, G),
\end{equation}

\noindent where $P$ represents the predicted segmentation map, and \(G\) denotes the ground truth mask for a given input image. The Dice loss \(\mathcal{L}_{\text{dice}}\) evaluates the overlap between the predicted map and ground truth, while the cross-entropy loss \(\mathcal{L}_{\text{ce}}\) measures pixel-wise classification accuracy.

\noindent The Dice loss is defined as:

\begin{equation}
\mathcal{L}_{\text{dice}}(P, G) = 1 - \frac{2 \sum_{j} P_j G_j}{\sum_{j} P_j^2 + \sum_{j} G_j^2 + \epsilon},
\end{equation}

\noindent where $P$ and \(G_j\) denote the predicted probability and ground truth value for the \(j\)-th voxel, respectively, and \(\epsilon\) is a small constant to ensure numerical stability.

\noindent The cross-entropy loss is defined as:

\begin{equation}
\mathcal{L}_{\text{ce}}(\hat{p}, G) = - \sum_{j} \big(G_j \log(P_j) + (1 - G_j) \log(1 - \hat{p}_j)\big),
\end{equation}

\noindent where the logarithmic terms account for the binary classification of each voxel.

\noindent \textbf{Rationale: Class Embedding Supervision Loss ($\mathcal{L}_{\text{ClassEmbed}}$)} In BiPVL-Seg, the vision encoder extracts generalized visual features at each stage, where the channel dimensions encode mixed information across all classes. By default, these embeddings have no explicit class-wise separation, meaning there is no guarantee that specific channels consistently correspond to specific anatomical structures or object classes. To address this, we introduce a lightweight MLP at the final encoder stage that maps the generalized feature channels into class-specific channels, ensuring each channel is dedicated to a particular class.

However, this mapping requires guidance to ensure consistency throughout training. Without supervision, the MLP could arbitrarily assign channels to different classes in each training iteration, disrupting alignment between encoder features and downstream segmentation predictions. To enforce consistent class assignment, we use the model’s own class logits from the vision decoder as pseudo-labels. These logits, which represent the decoder’s current class predictions, are downsampled to match the spatial resolution of the final encoder features. We then apply a Cross-Entropy loss between these class-specific encoder features and the pseudo-labels, ensuring that the encoder learns to produce class-separated features that align with the decoder’s segmentation outputs. This feedback loop progressively refines the encoder features, making them more semantically meaningful and directly linked to the segmentation task. As training progresses, the decoder’s predictions improve, providing better supervision signals for the encoder’s class-aware feature learning. This mechanism enhances the overall synergy between encoder and decoder, facilitating more consistent and interpretable feature representations within the vision encoder.

\noindent \textbf{Rationale: Alignment Loss ($\mathcal{L}_{\text{alignment}}$)} In BiPVL-Seg, a similarity matrix is constructed to measure how well the visual and textual embeddings align across all $N$ classes. Each row represents a visual class embedding, while each column represents a textual class embedding. Each entry in the matrix quantifies the similarity between a visual embedding from one class and a textual embedding from another class, enabling a comprehensive cross-modal comparison across all class pairs.

The diagonal elements of this matrix capture the similarity between visual and textual embeddings of the same class, representing the \textit{positive pairs}. The off-diagonal elements capture the similarity between embeddings of different classes, representing the \textit{negative pairs}. During contrastive learning, the objective is to maximize the similarity of positive pairs while minimizing the similarity of negative pairs. This optimization encourages vision and textual embeddings of the same class to converge in a shared multimodal space, while embeddings from different classes are pushed apart. By enforcing this structured separation, BiPVL-Seg enhances cross-modal alignment, ensuring that both visual and textual encoders contribute complementary information to the segmentation process.

\noindent\textbf{Balancing Multiple Loss Functions} 
BiPVL-Seg jointly optimizes three distinct tasks — segmentation (\(\mathcal{L}_{\text{seg}}\)), class embedding supervision (\(\mathcal{L}_{\text{ClassEmbed}}\)), and global-local alignment (\(\mathcal{L}_{\text{alignment}}\)). Each task operates at a different granularity and contributes complementary information for improving segmentation performance. However, these losses have inherently different magnitudes, convergence rates, and levels of supervision quality, making it impractical to assign static weights that remain optimal throughout training.

To address this, we adopt homoscedastic uncertainty-based weighting~\cite{kendall2018multi}, which dynamically balances the loss terms based on their inherent observation noise. Homoscedastic uncertainty captures task-specific noise that is constant across data points but varies between tasks. By modeling this uncertainty, BiPVL-Seg automatically learns the optimal contribution of each loss term, adjusting dynamically as training progresses.

The combined loss is presented in the form:
\begin{equation}
    \mathcal{L} = \beta_1\mathcal{L}_{\text{seg}} + \beta_2\mathcal{L}_{\text{ClassEmbed}} + \beta_3\mathcal{L}_{\text{alignment}},
\end{equation}
where \(\beta_1\), \(\beta_2\), and \(\beta_3\) are trainable weights. These weights are derived directly from the homoscedastic uncertainties, where:
\[
\beta_i = \frac{1}{2\sigma_i^2}
\]
for each task \(i\). This formulation ensures that tasks with higher uncertainty (indicating noisier or less reliable supervision) are automatically downweighted, while more reliable tasks contribute more to the total loss. The final combined objective is:

\begin{multline}
    \mathcal{L} = \frac{1}{2\sigma_1^2}\mathcal{L}_{\text{seg}} 
                + \frac{1}{2\sigma_2^2}\mathcal{L}_{\text{ClassEmbed}} 
                + \frac{1}{2\sigma_3^2}\mathcal{L}_{\text{alignment}} \\
                + \log \sigma_1 + \log \sigma_2 + \log \sigma_3,
\end{multline}

\noindent where the \(\log \sigma_i\) terms act as regularizers to prevent the uncertainties from collapsing to zero, ensuring stable optimization. This adaptive weighting mechanism allows BiPVL-Seg to balance the three tasks effectively in an end-to-end manner, removing the need for manual tuning and ensuring each component contributes proportionally to the overall objective.

\begin{figure}[t]
\centering
\includegraphics[width=0.8\columnwidth]{
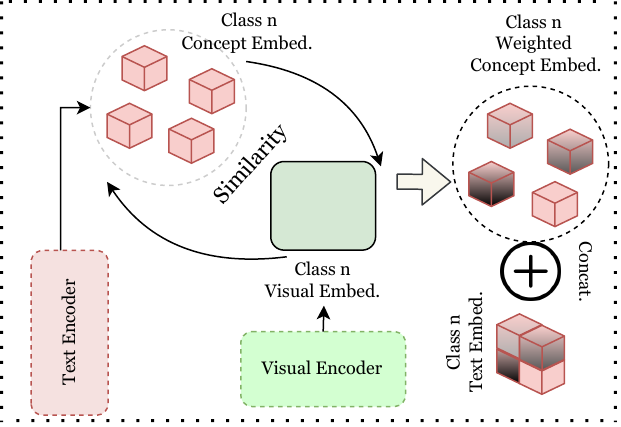}
\vspace{-10pt}
\caption{Text embeddings are created from concept embeddings. Different shades of concept embeddings emphasize that they have different weights.}
\vspace{-10pt}
% \vskip -0.2 in
\label{fig:text_embeddings}
\end{figure}

\noindent \textbf{Aggregated Text Embeddings} 
\Cref{fig:text_embeddings} illustrates how the aggregated text embeddings are constructed from individual concept embeddings using learnable weights. In the figure, lighter-colored concept embeddings gradually shift to different shades, indicating that their contributions are dynamically adjusted based on their similarity to the corresponding class’s visual embeddings. This adaptive weighting ensures that more visually relevant concepts contribute more strongly to the final class-level text embedding.

\noindent \textbf{Benchmark Vision-Only Models}

\begin{itemize}
    \item \textbf{U-Net}~\cite{ronneberger2015unet}: U-Net is a convolutional neural network (CNN) architecture specifically designed for biomedical image segmentation. Its encoder-decoder structure enables the extraction of hierarchical features, while skip connections ensure spatial details are preserved during upsampling. U-Net has become a go-to model due to its simplicity and effectiveness.

    \item \textbf{UNet++}~\cite{zhou2018unet++}: UNet++ extends U-Net by introducing dense skip pathways and nested architectures to improve feature propagation and gradient flow. These modifications enhance the segmentation accuracy, particularly for complex anatomical structures, while maintaining computational efficiency.

    \item \textbf{nnU-Net}~\cite{isensee2021nnu}: nnU-Net represents a highly automated U-Net framework that adapts its architecture and training pipeline to the dataset at hand. By incorporating automated preprocessing, augmentation, and hyperparameter tuning, nnU-Net consistently achieves state-of-the-art results across diverse medical imaging tasks.

    \item \textbf{STU-Net}~\cite{huang2023stu}: STU-Net integrates a Swin Transformer-based encoder into the U-Net structure, allowing it to capture both local and global features effectively. The hierarchical design of Swin Transformers improves segmentation performance, especially for volumetric medical images, by leveraging self-attention mechanisms to enhance contextual representation.

    \item \textbf{MedNeXt}~\cite{roy2023mednext}: MedNeXt introduces a scalable medical vision transformer architecture based on hierarchical depthwise convolutions and attention mechanisms. Unlike pure transformer-based models, MedNeXt balances computational efficiency with segmentation accuracy by combining convolutional inductive biases with attention-based global feature modeling.

    \item \textbf{UNETR}~\cite{hatamizadeh2022unetr}: UNETR employs a Vision Transformer as its encoder to capture global context, combined with a CNN-based decoder for high-resolution spatial predictions. This hybrid design enables the model to handle both global and local features effectively, making it well-suited for volumetric medical images.

    \item \textbf{Swin UNETR}~\cite{hatamizadeh2021swin}: Swin UNETR incorporates Swin Transformers into the encoder, leveraging their ability to process hierarchical features through shifted window attention. This approach improves segmentation performance by efficiently modeling long-range dependencies within medical images.

    \item \textbf{Swin UNETR-V2}~\cite{he2023swinunetr}: Building on Swin UNETR, Swin UNETR-V2 introduces further architectural enhancements such as refined attention mechanisms and optimized decoders. These improvements boost segmentation accuracy while maintaining computational efficiency, especially for complex 3D medical datasets.

    \item \textbf{nnFormer}~\cite{zhou2021nnformer}: nnFormer is a transformer-based segmentation model designed specifically for 3D medical imaging. It replaces convolutional encoders with a hierarchical transformer-based feature extraction mechanism, effectively modeling long-range spatial dependencies while maintaining fine-grained details. nnFormer achieves high segmentation accuracy across multiple medical imaging benchmarks.

    \item \textbf{MedFormer}~\cite{gao2022data}: MedFormer introduces a data-efficient transformer-based segmentation model that utilizes a locality-sensitive attention mechanism. This approach reduces the computational overhead of traditional transformers while preserving their ability to capture long-range dependencies. MedFormer is particularly effective in low-data scenarios, making it a strong candidate for medical image segmentation tasks with limited training samples.
\end{itemize}

\noindent \textbf{Benchmark Vision-Language Models} Here, below we introduce the vision-language models we utilized for comparison:
\begin{itemize}
    \item \textbf{Universal-CLIP}~\cite{liu2023clip}: Universal-CLIP enhances medical image segmentation by integrating CLIP-driven text embeddings, improving multi-modal alignment between visual and textual features. This approach enables robust multi-organ segmentation and tumor detection while ensuring strong generalizability across diverse medical datasets. We utilize Swin UNETR~\cite{he2023swinunetr} as the vision backbone, and the CLIP~\cite{radford2021learning} text encoder for our experiments. For the textual input, we follow their strategy to create the same format textual input.
    
    \item \textbf{MulModSeg}~\cite{li2024novel}: MulModSeg introduces a multi-modal medical image segmentation framework that integrates modality-conditioned text embeddings and an alternating training strategy. By leveraging a frozen CLIP text encoder, it enhances modality awareness without requiring significant architectural modifications. The alternating training procedure ensures balanced learning from unpaired CT and MR images, improving segmentation accuracy across both modalities. We utilize the similar settings as Universal-CLIP as these two models are very similar.

    \item \textbf{ZePT}~\cite{jiang2024zept}: ZePT introduces a zero-shot pan-tumor segmentation framework based on query-disentangling and self-prompting. It partitions object queries into fundamental and advanced subsets, first learning organ segmentation before refining tumor localization using self-generated visual and textual prompts. Additionally, ZePT incorporates feature-level query-knowledge alignment to enhance generalizability, enabling segmentation of unseen tumor categories without additional supervision. We utilize the default vision and text encoder for our experiments, and for textual input we use the detailed textual information by aggregating all of our concept descriptions into one single description per class.

    \item \textbf{CAT}~\cite{huang2024cat}: CAT introduces a dual-prompt segmentation model that coordinates anatomical and textual prompts for multi-organ and tumor segmentation. Anatomical prompts leverage 3D cropped images, while textual prompts incorporate medical domain knowledge to enhance segmentation accuracy. The model employs a query-based framework with a ShareRefiner module to refine segmentation and prompt queries. For CAT, we follow ZePT's settings as they are very similar to each other.
\end{itemize}

\section{Datasets, Metrics, and Additional Experiments}

\subsection{Datasets Description}
We selected diverse 3D imaging datasets to ensure comprehensive coverage across anatomical regions, class types, and imaging domains. 

\noindent\textbf{AMOS22}~\cite{ji2022amos} provides abdominal CT images with annotations for 15 organs: spleen, right kidney, left kidney, gallbladder, esophagus, liver, stomach, aorta, inferior vena cava, pancreas, right adrenal gland, left adrenal gland, duodenum, bladder, and prostate/uterus. The dataset consists of 200 images for training-validation and 100 for testing.  

\noindent\textbf{MM-WHS}~\cite{zhuang2019evaluation} includes whole-heart segmentation from CT and MR images, labeled across 7 cardiac structures: left ventricle blood cavity, myocardium of the left ventricle, right ventricle blood cavity, left atrium blood cavity, right atrium blood cavity, ascending aorta, and pulmonary artery. It comprises 20 CT and 20 MR images, evaluated separately as \textbf{MM-WHS (CT)} and \textbf{MM-WHS (MR)}.

\noindent\textbf{MSD-Brain}~\cite{antonelli2022medical} is part of the Medical Segmentation Decathlon, specifically focused on brain tumor segmentation. The dataset includes annotations for three tumor subregions: enhancing tumor (ET), tumor core (TC), and whole tumor (WT). For MSD-Brain, we generated multi-label segmentation to map the original labels to three tumor subregions: Tumor Core (TC), Whole Tumor (WT), and Enhancing Tumor (ET). Specifically, label 2 and label 3 were merged to form TC, labels 1, 2, and 3 were combined for WT, and label 2 was used for ET. This ensures a consistent multi-class segmentation representation aligned with prior brain tumor segmentation standards. From the publicly available images, we accessed 484 cases, splitting them into 349 for training, 39 for validation, and 96 for testing.

\begin{table}[t]
\centering

%\scriptsize
\small
\resizebox{0.8\columnwidth}{!}{
    \begin{tabular}{cc|cc}
        \toprule
        \multicolumn{2}{c|}{BiPVL-Seg Components}& \multicolumn{2}{c}{AMOS22 (CT)}\\
        \hline
        Vision&Language&\multirow{2}{*}{DSC$\uparrow$}&\multirow{2}{*}{NSD$\uparrow$}\\
        Backbone&Backbone&&\\
        \hline \hline
        Swin UNETR  & ClinicalBERT &88.21&93.97\\
        Swin UNETR  & BioBERT &87.91&93.19\\
        Swin UNETR  & PubMedBERT &88.01&93.82\\
        \hline
        UNet  & ClinicalBERT &83.79&90.36\\
        UNETR  & ClinicalBERT &79.88&87.45\\
        
        \bottomrule
    \end{tabular}
    }
    \caption{More ablation study on examining the effects of different vision and language backbones.}
    \label{tab:ablation3}
    % \vspace{-0.25in}
\end{table}

\subsection{Datasets Preprocessing} We implemented a comprehensive preprocessing pipeline to standardize and augment the 3D medical imaging datasets. The process began with cropping the regions of interest based on intensity thresholds to focus on the anatomical structures. Images were then reoriented to a consistent anatomical alignment in the RAS coordinate system. To enhance localization and balance class representation, we applied patch sampling guided by labeled regions, using fixed spatial dimensions (\(96 \times 96 \times 96\)). Data augmentation included random flipping along all three spatial axes and random rotations to account for spatial variability in the data. Additionally, intensity shifts were introduced probabilistically to simulate scanner variability and improve model robustness. These preprocessing steps ensured consistent input dimensions, increased diversity in the training data, and facilitated improved generalization of the segmentation models. We did some data-specific processing which is provided in \Cref{tab:dataset_processing}.

\noindent \textbf{Parameters Tuning}
\Cref{tab:parameter_tuning} provides a detailed overview of the parameter tuning process, including the key hyperparameters explored and their corresponding ranges.

\begin{table}[t]
\centering

%\scriptsize
\small
\resizebox{0.9\columnwidth}{!}{
    \begin{tabular}{c|c}
        \toprule
        Parameters&Values\\
        \hline \hline
        $\alpha_1$, $\alpha_2$&1,1\\
        $\beta_1$, $\beta_2$, $\beta_3$&learnable through training\\
        $\tau$&0.2\\
        Swin UNETR dimension space&48\\
        ClinicalBERT dimension space&768\\

        \bottomrule
    \end{tabular}
    }
    \caption{Key parameters and dimensions used in BiPVL-Seg, including loss weights, temperature, and encoder embedding sizes.}

    \label{tab:parameter_tuning}
    \vspace{-0.25in}
\end{table}

\subsection{Evaluation Metrics} We assess model performance using three primary metrics: Dice Similarity Coefficient (DSC, \%), 95\% percentile Hausdorff Distance (HD95, mm), and Normalized Surface Distance (NSD, \%)~\cite{nikolov2018deep}.

\noindent The Dice Similarity Coefficient (DSC) measures the overlap between the predicted segmentation map $P$ and the ground truth mask \(G\), defined as:

\begin{equation}
\text{DSC} = \frac{2 \sum_{j} P_j G_j}{\sum_{j} P_j^2 + \sum_{j} G_j^2} \times 100,
\end{equation}

\noindent where \(P_j\) and \(G_j\) denote the predicted probability and ground truth value for the \(j\)-th voxel. The DSC ranges from 0 to 100\%, with higher values indicating better segmentation quality.

\noindent The 95\% percentile Hausdorff Distance (HD95) evaluates the geometric alignment between the predicted segmentation boundary and the ground truth. It is defined as:

\begin{equation}
\text{HD95} = \operatorname{quantile}_{95\%} \big(\max_{x \in \partial \hat{p}} \min_{y \in \partial G} \|x - y\|\big),
\end{equation}

\noindent where \(\partial \hat{p}\) and \(\partial G\) represent the boundaries of the predicted segmentation map and ground truth mask, respectively, and \(\|\cdot\|\) denotes the Euclidean distance. HD95 measures the largest boundary error at the 95th percentile, with lower values indicating better geometric accuracy.

\noindent The Normalized Surface Distance (NSD) quantifies boundary alignment within a tolerance \(\tau\), defined as:

\begin{equation}
\text{NSD} = \frac{\lvert \{ x \in \partial \hat{p} : \min_{y \in \partial G} \|x - y\| \leq \tau \} \rvert}{\lvert \partial G \rvert} \times 100,
\end{equation}

\noindent where \(\lvert \cdot \rvert\) denotes the cardinality of the set, and \(\tau\) is a predefined tolerance threshold. NSD ranges from 0 to 100\%, with higher values indicating better surface alignment.

\subsection{More Ablation Studies}
\noindent\textbf{Different Textual Information Format} Below is one example of each of the different textual information formats we tried to evaluate our model:
\begin{itemize}
    \item Fixed, consistent concepts per class: ``Left Ventricle in a CT Image", ``Oval-like shape with thick walls tapering slightly towards the apex.", ``Positioned on the left side of the heart, below the left atrium, and anterior to the descending aorta.", ``High-density muscular wall visible due to contrast enhancement; the lumen appears lower in density when filled with blood or contrast agent.", ``Smooth and well-defined contour with symmetrical walls on both sides of the cavity.", ``Relatively homogenous muscular structure with striations sometimes visible in the myocardium."
    \item Non-fixed, ambiguous concepts per class: ``Some internal structure, possibly cardiac, seen in certain CT Images", ``Roughly rounded form, though its outline might stretch or compress based on various factors.", ``Displays areas of differing densities, but the boundaries between these areas could blur depending on contrast levels or scan angle.",  ``Mostly uniform interior, though subtle variations in texture or density may appear, especially under higher resolution."
    \item Single aggregated definition per class: ``Left Ventricle in a CT Image that has Oval-like shape with thick walls tapering slightly towards the apex. It is positioned on the left side of the heart, below the left atrium, and anterior to the descending aorta. It has high-density muscular wall visible due to contrast enhancement; the lumen appears lower in density when filled with blood or contrast agent. It is with smooth and well-defined contour with symmetrical walls on both sides of the cavity. It is relatively homogenous muscular structure with striations sometimes visible in the myocardium."
    \item Class names only, no concepts: ``Left Ventricle in a CT Image".
\end{itemize}
\noindent \textbf{Experimenting With Backbones}
We present some additional ablation studies, exploring different vision and language backbones, in  \Cref{tab:ablation3}. These results show that changing the text encoder has minimal impact on final performance, indicating that the hierarchical alignment mechanism effectively aligns textual and visual representations regardless of the initial text encoder weights. This suggests that BiPVL-Seg’s performance is not sensitive to the choice of language backbone, as the hierarchical alignment progressively harmonizes both modalities.

Additionally, replacing the vision backbone with alternatives such as UNet or UNETR leads to significant improvements over their original vision-only counterparts, demonstrating the effectiveness of BiPVL-Seg’s multimodal fusion and alignment strategy in enhancing segmentation performance across different encoder architectures.